\pdfoutput=1
\documentclass[11pt]{article}

\usepackage[final]{acl}

\usepackage{times}
\usepackage{latexsym}
\usepackage{latexsym}
\usepackage{makecell}
\usepackage{bbding}
\usepackage{graphicx}
\usepackage{subcaption} 
\usepackage{fontawesome}

\usepackage[T1]{fontenc}
\usepackage[utf8]{inputenc}

\usepackage{microtype}

\usepackage{inconsolata}

\usepackage{graphicx}

\title{Unifying AI Tutor Evaluation: An Evaluation Taxonomy for \\ Pedagogical Ability Assessment of LLM-Powered AI Tutors}

\author{
 \textbf{Kaushal Kumar Maurya},
 \textbf{KV Aditya Srivatsa},
 \\
 \textbf{Kseniia Petukhova} and 
 \textbf{Ekaterina Kochmar}
\\
Mohamed bin Zayed University of Artificial Intelligence
\\
 \small{
   \texttt{\{kaushal.maurya, vaibhav.kuchibhotla, kseniia.petukhova, ekaterina.kochmar\}@mbzuai.ac.ae}
 }
}

\begin{document}
\maketitle

%%%%%%%%%%%%%%%%%%%%%%%%%%%%%%%%%%%%%%%%%%%%%%%%%%%%%%%%%%%%%%%
\begin{abstract}
In this paper, we investigate \textit{whether current state-of-the-art large language models (LLMs) are effective as AI tutors and whether they demonstrate pedagogical abilities necessary for good AI tutoring in educational dialogues}. Previous efforts towards evaluation have been limited to subjective protocols and benchmarks. To bridge this gap, we propose a \textit{unified evaluation taxonomy} with eight pedagogical dimensions based on key learning sciences principles, which is designed to assess the pedagogical value of LLM-powered AI tutor responses grounded in student mistakes or confusions in the mathematical domain. We release {\tt MRBench} -- a new evaluation benchmark containing 192 conversations and 1,596 responses from seven state-of-the-art LLM-based and human tutors, providing gold annotations for eight pedagogical dimensions. We assess reliability of the popular \texttt{Prometheus2} and {\tt Llama-3.1-8B} LLMs as evaluators and analyze each tutor's pedagogical abilities, highlighting which LLMs are good tutors and which ones are more suitable as question-answering systems. We believe that the presented taxonomy, benchmark, and human-annotated labels will streamline the evaluation process and help track the progress in AI tutors' development.

\faGithub\ \href{https://github.com/kaushal0494/UnifyingAITutorEvaluation}{\url{https://github.com/kaushal0494/UnifyingAITutorEvaluation}}

\end{abstract}

\section{Introduction}
\label{sec:intro}
Human tutoring is a cornerstone of educational development, playing a crucial role in fostering societal growth by empowering learners. While one-on-one tutoring is highly effective \cite{bloom19842}, its ubiquitous implementation is hindered by the limited availability of qualified tutors.\footnote{\url{https://unesdoc.unesco.org/ark:/48223/pf0000385723}} The remarkable success of LLMs as conversational systems offers promising opportunities in education \cite{wang2024large, gan2023large}, driving the development of LLM-powered intelligent tutoring systems (ITS) \citep{pal2024autotutor, liu2024teaching} and the deployment of LLMs as tutors using advanced prompting techniques \cite{denny2024generative, mollick2024instructors}. Such AI tutors serve various educational objectives~\cite{wollny2021we}, among which the task of students' mistake and confusion remediation is one of the most popular, leading to active AI tutor development \cite{macina2023mathdial, wang2024bridging}.

\begin{table}[!t]
    \centering
    \resizebox{0.45\textwidth}{!}{
    \begin{tabular}{|l|c|c|c|c|c|}
     \hline
        \textbf{Dimension} & \textbf{TP'22} & \textbf{MA'23} & \textbf{WA'24}  & \textbf{DA'24} & \textbf{Ours} \\ \hline \hline
        Mistake identification & \textcolor{green}\Checkmark & \textcolor{green}\Checkmark & \textcolor{red}\XSolid & \textcolor{green}\Checkmark & \textcolor{green}\Checkmark \\ \hline
        Mistake location & \textcolor{red}\XSolid & \textcolor{red}\XSolid & \textcolor{red}\XSolid & \textcolor{green}\Checkmark & \textcolor{green}\Checkmark \\ \hline
        Revealing of the answer & \textcolor{red}\XSolid & \textcolor{green}\Checkmark & \textcolor{red}\XSolid & \textcolor{red}\XSolid & \textcolor{green}\Checkmark \\ \hline
        Providing guidance & \textcolor{green}\Checkmark & \textcolor{red}\XSolid & \textcolor{green}\Checkmark & \textcolor{red}\XSolid & \textcolor{green}\Checkmark \\ \hline
        Actionability & \textcolor{red}\XSolid & \textcolor{red}\XSolid & \textcolor{red}\XSolid & \textcolor{green}\Checkmark & \textcolor{green}\Checkmark \\ \hline
        Coherence & \textcolor{red}\XSolid & \textcolor{green}\Checkmark & \textcolor{red}\XSolid & \textcolor{red}\XSolid & \textcolor{green}\Checkmark \\ \hline
        Tutor tone & \textcolor{green}\Checkmark & \textcolor{red}\XSolid & \textcolor{green}\Checkmark & \textcolor{red}\XSolid & \textcolor{green}\Checkmark \\ \hline
        Human-likeness & \textcolor{green}\Checkmark & \textcolor{red}\XSolid & \textcolor{green}\Checkmark & \textcolor{red}\XSolid & \textcolor{green}\Checkmark \\ \hline
    \end{tabular}}
    \caption{\small Evaluation dimensions considered in previous research on  AI tutoring for student mistake remediation. \textit{TP'22} stands for \citet{DBLP:conf/edm/TackP22}, \textit{MA'23} –  \citet{macina2023mathdial}, \textit{WA'24} – \citet{wang2024bridging}, and \textit{DA'24} – \citet{daheim2024stepwise}.}
    \vspace{-0.5cm}
    \label{tab:comapp}
\end{table}

While the development of AI tutoring systems presents significant challenges, evaluating their pedagogical abilities is even more challenging and crucial for tracking the efficacy and quality of AI tutoring. General domain-agnostic natural language generation (NLG) metrics~\citep{lin2004rouge, popovic2017chrf++, post2018call, gao2020dialogue, liu2023g} are not well-suited for this context, as most of them fail to account for pedagogical values and require gold references, which are often not available, especially in online interactions. Specifically, for the student mistake remediation task, we need to assess complex pedagogical aspects and abilities of such systems, ensuring that they provide students with sufficient, helpful, and factually correct guidance and do not simply reveal answers when a student makes a mistake. For instance, \citet{macina2023mathdial} found that {\tt ChatGPT} as a tutor \textit{reveals the solution 66\% of the time} and provides \textit{incorrect feedback 59\% of the time}.

Despite recent efforts to incorporate pedagogical dimensions in the evaluation of AI tutoring systems, there is a notable lack of a unified evaluation taxonomy. For example, ~\citet{DBLP:conf/edm/TackP22} and \citet{tack2023bea} evaluated models' responses from the perspective of whether they {\em speak like a teacher}, {\em understand a student}, and {\em help a student}. ~\citet{macina2023mathdial} assessed the responses of models acting as tutors focusing on {\em coherence}, {\em correctness}, and {\em equitable tutoring}. Finally,~\citet{wang2024bridging} evaluated {\em usefulness}, {\em care}, and {\em humanness}, while \citet{daheim2024stepwise} focused on {\em targetedness}, {\em correctness}, and {\em actionability} to assess the quality of tutor responses. Table \ref{tab:comapp} presents an overview of these approaches. The disparity in the evaluation schemata and definitions used and lack of standardization pose significant challenges in tracking the progress and actual performance of existing AI tutors and complicate the comparison between different systems. Moreover, existing taxonomies are often too abstract \cite{DBLP:conf/edm/TackP22}, compress multiple dimensions into a single criterion \cite{daheim2024stepwise}, or are incomplete \cite{wang2024bridging, macina2023mathdial}.

To address these issues, we propose \textbf{the first unified evaluation taxonomy based on learning sciences principles} to assess the pedagogical abilities of AI tutors. This  taxonomy is centered around eight evaluation dimensions related to \textit{student mistake remediation} including: (1) \textit{mistake identification}, (2) \textit{mistake location}, (3) \textit{revealing of the answer}, (4) \textit{providing guidance}, (5) \textit{actionability}, (6) \textit{coherence}, (7) \textit{tutor tone}, and (8) \textit{human-likeness}. As we elaborate in Section \ref{sec:evaltax}, our taxonomy is strongly aligned with key pedagogical values and unifies the taxonomies used in previous research.

In addition to the taxonomy, we compile and release {\tt MRBench} – {\bf a new evaluation benchmark} derived from two public datasets, \texttt{MathDial} \cite{macina2023mathdial} and {\tt Bridge} \cite{wang2024bridging}. Each instance in the benchmark includes a partial conversation between a tutor and a student, concluding when the student either makes a mistake or exhibits confusion. The instance is also associated with the following human tutor's response aimed at remediating the student's mistake. Using these partial conversation histories exhibiting students' mistakes, we generated responses from seven state-of-the-art LLMs acting as tutors and conducted human and LLM-based evaluations to assess the pedagogical abilities of these models. Our findings indicate that while state-of-the-art LLMs like {\tt GPT-4} are effective question-answering systems, they are often not as competent as tutors. In summary, our key contributions are as follows:

\begin{itemize} \vspace{-0.5em}
    \item We present a unified evaluation taxonomy with eight dimensions to assess the pedagogical abilities of LLM-based AI tutors. Grounded in the learning sciences principles, this taxonomy evaluates the effectiveness of AI tutors for student mistake remediation within the mathematics domain.
    \vspace{-0.5em}
    \item We release {\tt MRBench}, an evaluation benchmark based on existing datasets and containing responses from 7 state-of-the-art LLMs acting as tutors, which are annotated using the proposed taxonomy.
    \vspace{-0.5em}
    \item We investigate the pedagogical abilities of LLMs as AI tutors via human and LLM-based evaluation. Additionally, we discuss the reliability of LLM-based evaluation by correlating it with human judgements.
    \vspace{-0.5em}
    \item The taxonomy, benchmark, and human annotations will be made publicly available to facilitate future research in this important domain.
\end{itemize}

\section{Related Work}
\label{sec:relwork} 
In this section, we first briefly overview and discuss the limitations of the existing general-purpose NLG metrics and then turn to pedagogically-oriented approaches to evaluation. 

\subsection{General NLG and LLM-based Evaluation} \label{sec:nlgmet}
General domain-agnostic natural language generation (NLG) metrics like BLEU \cite{papineni2002bleu}, BERTScore \cite{lin2004rouge}, DialogRPT \cite{gao2020dialogue}, and so on have been used as proxies to measure the coherence and human-likeness of AI tutor responses. However, these metrics do not account for pedagogical values \cite{jurenka2024towards, liu2024teaching} and often require a ground truth answer to evaluate matching responses. For a given input dialogue, there can be multiple valid, pedagogically correct ground truth responses, making detection of the \textit{optimal} answer non-deterministic \cite{DBLP:conf/edm/TackP22, al2024can}. Additionally, these metrics can be easily manipulated; for instance, simple responses like ``Hello'' or ``teacher:'' \cite{baladon2023retuyt,jurenka2024towards} can inflate scores. While nowadays LLMs are used for AI tutor evaluation \cite{chevalier2024language} with respect to the model's helpfulness and human-likeness, among other aspects, their judgements are often unreliable \cite{wang2023large}.

\subsection{Pedagogically-oriented Evaluation}
\label{sec:pedaeval}
Most of the traditional evaluation methods from learning sciences are designed for the evaluation of human tutors and can not be easily applied to AI tutors due to the absence of a \textit{self-reports} \cite{DBLP:conf/edm/TackP22}. A reliable avenue is to hire human experts to evaluate pedagogical performance \cite{vasselli2023naisteacher, lee2023learning, abdelghani2024gpt}. However, there is no agreed-upon protocol for conducting pedagogical human evaluations, and researchers consider various pedagogical dimensions and their associated definitions \cite{wollny2021we, tack2023bea, borges-etal-2024-teach, denny2024generative}. The most commonly used evaluation framework involves human raters comparing the responses of two tutors in the context of the same dialogue snippet \cite{DBLP:conf/edm/TackP22}. These comparisons are based on the three dimensions defined by \citet{demszky2021measuring}, but they do not fully capture pedagogical richness. Similar efforts by \citet{macina2023mathdial, wang2024bridging, daheim2024stepwise} partially cover pedagogical aspects in the student mistake remediation task in the mathematical domain. A large-scale study by \citet{jurenka2024towards} concluded that there is a need to develop well-recognized, unified evaluation metrics that enable comparisons across different models and track the progress of AI tutors. Our proposed taxonomy is a step toward this ambitious goal, and we believe this effort will streamline the evaluation of AI tutors and their pedagogical abilities.

\section{Student Mistake Remediation Task}
\label{sec:stumisrem}
In this work, we focus on educational dialogues between a student and a tutor in the mathematical domain. Specifically, the conversations are grounded in students' mistakes or confusions, and the AI tutor aims to respond in order to remediate such mistakes or confusions. 

Formally, let's define the conversation history between a tutor and a student as  \( \mathcal{H} = \{ (\mathcal{T}_1, \mathcal{S}_1), (\mathcal{T}_2, \mathcal{S}_2), \ldots, (\mathcal{T}_t, \mathcal{S}_t) \}, \) where \(\mathcal{T}_i\) represents the \(i\)-th response from the tutor, and \(\mathcal{S}_i\) represents the \(i\)-th response from the student. Let \(\mathcal{S}_k\) denote the student's \textit{most recent} $k$ utterances, where $k \in [1, ..., t]$, containing a mistake or confusion. Then the objective of the tutor is to provide the most appropriate response \(\mathcal{T}_{t+1}\) to address this mistake or confusion. The evaluation taxonomy detailed in Section \ref{sec:evaltax} assesses the appropriateness of the $\mathcal{T}_{t+1} $ response across eight key pedagogical dimensions.

\begin{table*}[!t]
    \centering
    \resizebox{\textwidth}{!}{
    \begin{tabular}{|l|l|c|}
         \hline
         \textbf{Dimension} &  \textbf{Definition} & \textbf{Desiderata} \\ \hline
        Mistake identification & Has the tutor identified/recognized a mistake in a student’s response? & Yes \\ \hline
        Mistake location & Does the tutor’s response accurately point to a genuine mistake and its location? & Yes \\ \hline
        Revealing of the answer & Does the tutor reveal the final answer (whether correct or not)? &  No\\ \hline
        Providing guidance & \makecell[l]{Does the tutor offer correct and relevant guidance, such as an explanation, \\ elaboration, hint, examples, and so on?} & Yes \\ \hline
        Actionability & Is it clear from the tutor’s feedback what the student should do next? & Yes \\ \hline
        Coherence & Is the tutor’s response logically consistent with the student’s previous responses? & Yes \\ \hline
        Tutor tone & Is the tutor’s response encouraging, neutral, or offensive? & Encouraging \\ \hline
        Human-likeness & Does the tutor’s response sound natural rather than robotic or artificial? & Yes \\ \hline
    \end{tabular}}
    \caption{\small An overview of the proposed evaluation taxonomy.}
    \vspace{-0.3cm}
    \label{tab:brief_def}
\end{table*}

\section{Evaluation Taxonomy} 
\label{sec:evaltax}
In this section, we first present our approach, narrowing the evaluation taxonomy down to eight measurable dimensions aligned with key pedagogical strategies \cite{jurenka2024towards, hennessy2016developing}. These dimensions are most suitable for the student mistake remediation task and are based on the learning sciences principles. We then dive into the details of each dimension and its relationship to previous research. An overview of the taxonomy is presented in Table \ref{tab:brief_def}.

\paragraph{Grounding the Taxonomy in the Learning Sciences Principles} 
Considering tutors as expert advisors, we prioritize the following high-level pedagogical principles:
\begin{enumerate}
    \itemsep0em
    \item \textbf{Encourage active learning} \cite{chi2014icap, oakley2021uncommon}: The tutor should encourage students to actively participate in the  discussion and practice rather than passively receive information. The tutor can achieve this by \textit{not revealing the answer immediately} and \textit{scaffolding guidance}.

    \item \textbf{Adapt to students’ goals and needs} \cite{king2017reimagining}: The tutor should respond coherently by adapting to the current state and goals of the student's learning rather than following a pre-defined learning path. In the context of student mistake remediation, this happens when the tutor \textit{identifies the mistake}, \textit{pinpoints its location}, and \textit{responds coherently}.

    \item \textbf{Manage cognitive load and enhance metacognitive skills} \cite{mayer2002multimedia, dehaene2020we, cohen2021metacognitive}: The tutor should present the information in a structured manner, with elaboration and examples in manageably small chunks that enable the student to generalize their learning skills beyond the current problem. For the task at hand, this can be achieved by \textit{providing appropriate guidance}.

    \item \textbf{Foster motivation and stimulate curiosity} \cite{keller1987development, patall2008effects}: The tutor should constantly motivate and stimulate curiosity in the student throughout the dialogue, as this leads to self-efficacy and lifelong learning. For student mistake remediation, this can be achieved by clearly providing the \textit{next actionable step}, using an \textit{encouraging tone}, and behaving like a \textit{human expert tutor}.
\end{enumerate}

\subsection{Evaluation Taxonomy Dimensions} 
\label{sec:dimevaltax} 
This section delineates the specifics of each dimension of our taxonomy and elucidates its relationship to existing research.

\begin{enumerate}
    \itemsep0em
    \item {\bf Mistake identification}: Since all dialogues in the dataset contain a mistake made by the student, a good-quality response from the tutor should include the relevant mistake identification. This corresponds to {\em student understanding} in the schema of \citet{DBLP:conf/edm/TackP22} and {\em correctness} in the schemata of \citet{macina2023mathdial} and \citet{daheim2024stepwise}.
    
    \item {\bf Mistake location}: A good tutor response should not only notify the student of the committed error but also point to its location in the answer and outline what the error is to help the student remediate it in their subsequent response. This corresponds to {\em targetedness} in \citet{daheim2024stepwise}.
    
    \item {\bf Revealing of the answer}: Since most dialogues are relatively short and present contexts for the mistakes made early in the student's solution, a good tutor strategy is not to reveal the answer to the student immediately but rather provide helpful guidance. This aspect corresponds to {\em equitable tutoring} in \citet{macina2023mathdial}.
    
    \item {\bf Providing guidance}: In addition to not revealing the answer immediately, a good tutor response should provide the student with relevant and helpful guidance, such as a hint, an explanation, or a supporting question. This aspect corresponds to {\em helping a student} in \citet{DBLP:conf/edm/TackP22} and {\em usefulness} in \citet{wang2024bridging}.
    
    \item {\bf Actionability}: Once the guidance is provided to a student, it should be clear from a good tutor response what the student should do next; in other words, the tutor response should not be vague, unclear, or a conversation stopper. This aspect in our schema corresponds to {\em actionability} in
    \citet{daheim2024stepwise}.

    \item \textbf{Coherence}: We postulate that a high-quality tutor's response should be \textit{logically consistent} with the student's previous responses. This aligns with the \textit{coherence} aspect from \citet{macina2023mathdial}.

    \item \textbf{Tutor tone}: In addition to addressing student mistakes, a good tutor should encourage them and avoid using toxic language, which is aligned with the \textit{care} dimension in the evaluation schema of \citet{wang2024bridging}. This dimension is particularly critical for LLM-based AI tutors, as they often exhibit unpredictable behavior.

    \item \textbf{Human-likeness}: Effective tutoring requires that students feel a connection with the tutor, which is more likely when the tutor's responses appear human-like rather than robotic. This aspect corresponds to the \textit{human-likeness} dimension in \citet{wang2024bridging}'s schema.
\end{enumerate}

Overall, our schema covers all the relevant aspects of a good tutor response proposed in previous work~\cite{DBLP:conf/edm/TackP22,macina2023mathdial,wang2024bridging,daheim2024stepwise} while also being supported by the learning sciences principles. Although there are inherent inter-dependencies among the proposed dimensions of the taxonomy (e.g., a response that reveals the answer is less likely to be actionable, and vice versa), we explicitly instructed all annotators to treat each dimension as \textit{independent} and \textit{orthogonal} to minimize confounding factors and potential biases during the annotation process. Estimation of the relative importance among evaluation dimensions is beyond the scope of this study and is left for future work. 

\subsection{Evaluation Taxonomy Validation} 
\label{sec:valideevaltax}
To evaluate the efficacy and pertinence of the proposed evaluation taxonomy, we conducted a series of validation experiments aimed at addressing the critical questions: \textit{Are the proposed dimensions sufficient?} and {\em Are there redundancies among them?} The annotation team consisted of two male and two female annotators, with all four annotators holding at least a post-graduate degree in Computer Science and being proficient in English. We note that for this study, we do not require annotators to have direct teaching experience, as understanding of the mathematical tasks at the middle school level and being able to judge the responses from the perspective of a potential user of such AI tutors (or a student), rather than specifically a teacher, is sufficient. To control the annotation workflow and ensure quality, we opted not to use public annotation outsourcing platforms such as Prolific or MTurk, which allowed us to implement rigorous training protocols and a robust validation mechanism for the annotations.

First, we provided all annotators with comprehensive training, including an interactive training document (see Section \ref{appsec:humann} for more details) and oral instructions. Following this, we conducted \textit{validation pilot study} to evaluate the annotation quality and the annotators' understanding of the instructions before rolling out the large-scale human evaluation detailed in Section \ref{sec:humananno}. This multi-step process ensured that the annotations adhered to our quality standards. In this \textit{validation pilot study}, all four annotators iteratively reviewed the annotation scheme and guidelines. Each annotator also independently labeled the same eight randomly sampled dialogues -- four from each of the two datasets ({\tt Bridge} and \texttt{MathDial}) -- across the eight dimensions of the evaluation taxonomy. Given that each dialogue contained multiple responses from both LLMs and humans, and each response was annotated across eight evaluation dimensions, this resulted in a total of \textbf{544 annotations per annotator}. To measure inter-annotator agreement, we computed Fleiss' kappa value, which for this annotation experiment equals 0.65, indicating substantial agreement. None of the annotators identified any additional or redundant dimensions necessary for student mistake remediation.

\section{Pedagogical Ability Assessment Settings} 
\label{sec:expsetup}
In this section, we provide details on the benchmark data preparation and statistics, LLMs deployed as AI tutors, the human annotation process, and the LLM-based evaluation.

\subsection{Benchmark Preparation}
\label{sec:benchprep}
We have compiled mistake remediation benchmark, {\tt MRBench}, from the {\tt Bridge} \cite{wang2024bridging} and \texttt{MathDial} \cite{macina2023mathdial} datasets. Each instance in both datasets comprises educational dialogue interactions between students and tutors within the mathematical domain. These interactions are specifically anchored in the students' errors or misconceptions, accompanied by the subsequent human tutor response, which aims to remediate the mistake or confusion.

The {\tt Bridge} dataset~\cite{wang2024bridging} comprises partial dialogue interactions between real human tutors and students at the elementary level, featuring two distinct human tutor responses (novice and expert). The dialogue context is typically short (few turns) and predominantly focused on fundamental mathematical concepts, including operations such as multiplication, addition, and so on. The original dataset consists of a total of 700 dialogues; we filtered 60 high-quality instances for {\tt MRBench}.\footnote{Many examples in the dataset are grounded in visual contexts, which we have not incorporated into this benchmark.} Among the various criteria for selecting high-quality dialogues, the key one was that the student's last utterance (or last few utterances) should exhibit an error or confusion.

The dialogues in the \texttt{MathDial} dataset~\cite{macina2023mathdial} consist of complete multi-turn conversations between a real human tutor and an LLM acting as a student, where the tutor aims to remediate the student's mistakes. Specifically, these conversations are grounded in middle school-level mathematical reasoning questions. To match the format of {\tt Bridge} (partial conversations with the last few student's utterances exhibiting a mistake or confusion), we prepared the dataset by terminating a conversation where the student makes a mistake and considering the next tutor response as the expert tutor response (there are no associated novice responses in this dataset). To further ensure the reliability of our benchmark, we manually inspected the data in order to retain only high-quality examples, which resulted in 132 instances for {\tt MRBench}.

Next, for the 192 instances in {\tt MRBench} (60 from {\tt Bridge} and 132 from \texttt{MathDial}), we generated appropriate subsequent responses based on the conversation history and the last utterance, which contained confusions or mistakes, using seven state-of-the-art LLMs. These models were prompted to act as expert tutors (see Figure \ref{fig:promptresponse}  for the exact prompt template). We consider state-of-the-art LLMs of various sizes and capabilities, including: {\tt GPT-4} \cite{achiam2023gpt}, {\tt Gemini} \cite{reid2024gemini}, {\tt Sonnet} \cite{TheC3}, {\tt Mistral} \cite{jiang2023mistral}, {\tt Llama-3.1-8B} and {\tt Llama-3.1-405B} \cite{dubey2024llama}, and {\tt Phi3}~\cite{abdin2024phi}.

The diverse formats of the two datasets in our benchmark, with varying difficulty levels, make it more suitable for assessing the pedagogical abilities of the AI tutors in different scenarios. Furthermore, each LLM has associated responses for 192 dialogues, resulting in a benchmark of 192 × 7 (7 LLM responses) + 192 × 1 (expert responses) + 60 × 1 (novice responses) = 1,596 responses, which makes the evaluation benchmark reasonably large while still manageable for human annotation described in Section \ref{sec:humananno}. More details on benchmark statistics are presented in Appendix Section \ref{appsec:benchstats}.

\subsection{Human Annotation} 
\label{sec:humananno}

\begin{figure}
    \centering
    \includegraphics[width=1\linewidth]{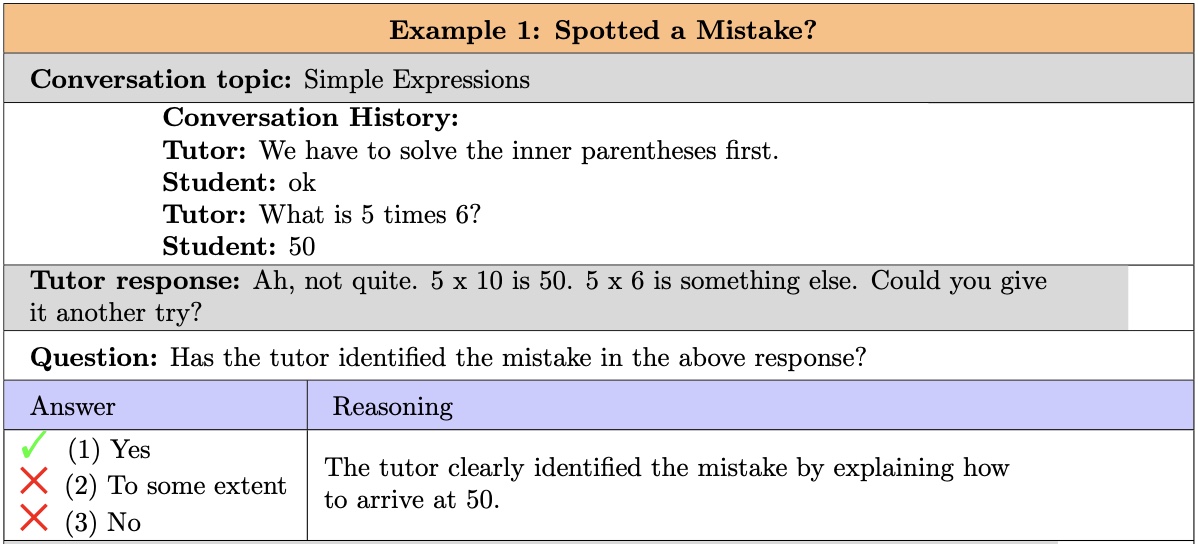}
    \vspace{-0.3cm}
    \caption{An example of mistake identification from the validation pilot study.}
    \vspace{-0.5cm}
    \label{fig:example}
\end{figure}

Four trained annotators (see Section \ref{sec:valideevaltax}) annotated {\tt MRBench} using the validated taxonomy. Each annotator was asked to annotate human and LLM-based tutor responses across 8 dimensions of the taxonomy in the context of 48 dialogues. A total of 192 instances were annotated, with 40 of those annotated independently by two annotators (10 instances from {\tt Bridge} and 30 from \texttt{MathDial}) allowing us to calculate pairwise inter-annotator agreement. Each dimension was annotated using a three-tier labeling system (see Figure \ref{fig:example} and Table \ref{tab:det_def}). For instance, the `mistake identification' dimension employed the following labels: (i) yes, (ii) to some extent, and (iii) no. Annotators were instructed to assign `yes' if the tutor accurately identified the mistake, `no' if the mistake was missed, and `to some extent' when there was ambiguity or uncertainty in the mistake identification. The annotators reached an average Cohen’s kappa score of $0.71$, which indicates substantial inter-annotator agreement~\cite{mchugh2012interrater}. 

\begin{table*}[!t]
    \centering
    \resizebox{\textwidth}{!}{ 
    \begin{tabular}{@{}lcccccccc@{}}
        \hline \hline
        \textbf{Tutor}      & \textbf{Mistake Identification} & \textbf{Mistake Location} & \textbf{Revealing of the Answer} & \textbf{Providing Guidance} & \textbf{Actionability} & \textbf{Coherence}  & \textbf{Tutor Tone} & \textbf{Human-likeness} \\ \hline \hline
        \texttt{*Novice}      & 43.33                 & 16.67            & 80.00                  & 11.67              & 1.67     &  50.00         & 90.00     & 35.00     \\
        \texttt{Expert}      & 76.04                 & 63.02            & 90.62                  & 67.19              & \textbf{76.04}    &   79.17       & 92.19     & 87.50     \\ \hline
        
        {\tt Llama-3.1-8B}   & 80.21                 & 54.69            & 73.96                  & 45.31              & 42.71    & 80.73         & 19.79     & 93.75     \\
        {\tt Phi3}      & 28.65                 & 26.04            & 73.96                  & 17.71              & 11.98    & 39.58         & 45.31     & 52.08     \\ \hline 
        {\tt Gemini}      & 63.02                 & 39.58            & 67.71                  & 37.50              & 42.71    &  56.77        & 21.88     & 68.23     \\
        {\tt Sonnet}     & 85.42                 & 69.79            & \textbf{94.79}                  & 59.38              & 60.94    & 88.54         & \textbf{54.69}     & \textbf{96.35}     \\
        {\tt Mistral}     & 93.23                 & 73.44            & 86.46                  & 63.54              & 70.31    & 86.98         & 15.10     & 95.31     \\ \hline
        {\tt GPT-4}       & \textbf{94.27}              & \textbf{84.38}            & 53.12                  & 76.04              & 46.35    & 90.17         & 37.50     & 89.62     \\
        {\tt Llama-3.1-405B} & \textbf{94.27}                 & \textbf{84.38}            & 80.73                  & \textbf{77.08}              & 74.48    & \textbf{91.67}         & 16.15     & 90.62     \\ \hline
        \hline
    \end{tabular}}
    \vspace{-0.8mm}
    \caption{Pedagogical ability assessment of different LLMs using the DAMR scores (in \%) across eight evaluation dimensions with \textit{human evaluation} on {\tt MRBench}. *For the {\tt Novice}, we have considered only 60 dialogues from the {\tt Bridge} dataset. The DAMR scores for  {\tt Novice} are reported on these 60 instances, while for {\tt Expert} and all LLMs, all 192 instances were considered. The best DAMR scores for each dimension are \textbf{bolded}.}
    \label{tab:main_results}
    \vspace{-0.8mm}
\end{table*}

\subsection{LLM-based Annotation} 
\label{sec:llmanno}

Due to the growing interest in utilizing LLMs as critics or evaluators \cite{jurenka2024towards, 10.1145/3641289}, we also used two LLMs as evaluators: 
\begin{itemize}
    \item We used {\tt Prometheus2} \cite{kim2024prometheus} because: (i) it was specifically trained as an evaluator using reinforcement learning with human feedback (RLHF), (ii) it has a high correlation with human annotations and \texttt{GPT-4}, and (iii) it does not belong to any of the LLM families considered as AI tutors in our framework.
    \item In addition, we also used {\tt Llama-3.1-8B} as a lightweight LLM to assess the reliability of smaller models that were not fine-tuned for evaluation objectives as a critic. 
\end{itemize}

\subsection{Assessment Metrics}
\label{sec:metrics}
We utilize two key metrics to quantitatively assess the pedagogical effectiveness of LLMs and for comparative analysis: (1) \textbf{Desired Annotation Match Rate (DAMR)}: This metric quantifies the percentage of responses from each human or LLM-based tutor that received the \textit{desired} annotation labels. The \textit{desired} labels for each dimension are detailed in Table \ref{tab:brief_def}. This metric offers a comparative analysis of response quality across human tutors and various LLMs, providing insights into their pedagogical performance. (2) \textbf{Annotation Correlation (AC)}: This metric is based on \textit{Pearson’s correlation} \cite{sedgwick2012pearson}, and it estimates the correlation between LLM-generated and human annotations \cite{kim2024prometheus}, allowing us to assess the reliability of LLMs as evaluators in the context of student mistake remediation.

\section{Key Findings}
\label{sec:result&discc}
This section summarizes the key findings of our study on the pedagogical abilities of LLMs as AI tutors, based on human and LLM-based evaluation of {\tt MRBench}, and the correlation between them. We consider human-based evaluations as gold standard. Table \ref{tab:main_results} shows DAMR scores for each LLM across all eight dimensions.

\paragraph{Performance of the powerful {\tt \textbf{GPT-4}} and {\tt \textbf{Llama-3.1-405B}} models:} Both these LLMs perform well in identifying students' mistakes and their exact location, with {\tt Llama-3.1-405B} having a slight edge as {\tt GPT-4} reveals the answer approximately 47\% of the time, making its responses less actionable and impacting student's learning experience. \textit{This shows that {\tt GPT-4} is a good question-answering system but a relatively poor tutor.} At the same time, {\tt GPT-4}'s responses tend to be more encouraging. The guidance score is also high because {\tt GPT-4}'s answer-revealing responses often offer useful explanations, providing the student with learning opportunities. {\tt Llama-3.1-405B} performs more robustly along these dimensions, though it is less encouraging. Both models exhibit a high level of coherence, and their responses are human-like as indicated by high DAMR scores.

\paragraph{Performance of {\tt \textbf{Gemini}}, {\tt \textbf{Sonnet}}, and {\tt \textbf{Mistral}}:} Among these three LLMs, \texttt{Gemini} performs the worst as its responses are often incoherent, while also achieving low scores for mistake identification and exact location. Furthermore, even in coherent responses, the model frequently reveals the answer and receives low scores for actionability and guidance as its explanations for both correct and incorrect revealed answers are often factually inaccurate, harming students' learning. {\tt Sonnet} and {\tt Mistral} perform slightly better than {\tt Gemini}, with {\tt Sonnet} focusing primarily on encouraging tone and human-likeness while avoiding revealing the answer, though it is less effective along key pedagogical dimensions like mistake identification, location, guidance, and actionability. On the other hand, {\tt Mistral} shows a slight edge along each of the dimensions. In conclusion, among these three models, {\tt Gemini} performs the worst, and {\tt Mistral} performs slightly better than {\tt Sonnet}.

\paragraph{Performance of {\tt \textbf{Llama-3.1-8B}} and {\tt \textbf{Phi3}}:} To account for diversity, we also included two lightweight LLMs (with fewer parameters) as tutors, namely {\tt Llama-3.1-8B} and {\tt Phi3}. {\tt Phi3} is the worst-performing LLM model in this context, with the lowest score for coherence, suggesting that the responses from {\tt Phi3} are often irrelevant to the conversation context, as well as overall low scores in other dimensions. This underscores the model's inadequate capacity for contextual understanding and semantic alignment in educational dialogues considered in this study. In the few cases where {\tt Phi3} demonstrates some competence, it frequently reveals the answer, reflecting more of a question-answer system than a pedagogical tutor behavior. Moreover, its outputs tend to be robotic, template-based and lack the nuance expected in human responses. In contrast, despite having fewer parameters, \texttt{Llama-3.1-8B} demonstrates reasonable performance, albeit still below that of larger LLMs. Specifically, its responses are coherent, strategically avoid immediate answer revelation, robustly identify and rectify mistakes, and exhibit human-like behavior, as evidenced by the DAMR scores.

\paragraph{{\tt \textbf{Novice}} and {\tt \textbf{Expert}} human responses:} We also investigated the pedagogical value of human responses for both {\tt Novice} and {\tt Expert}. It can be observed that {\tt Novice} responses do not have a high score for guidance and are poor in terms of actionability (DAMR score of 1.67). Furthermore, the responses are generally short and ambiguous, such as "this is a good try," which leads to lower scores for mistake identification and location. At the same time, they often do not reveal the answer. In contrast, {\tt Expert} human responses are more logical and highly actionable for the next steps. However, in a few cases, their responses are action-oriented even though they do not provide factually correct guidance, leading to lower scores for guidance compared to actionability. This leads to a question: \textit{Can a tutor achieve a higher DAMR score for actionability while receiving a lower score for providing guidance?} This is possible since we consider only factually correct guidance as useful (see Table \ref{tab:det_def}). At the same time, even incorrect or incomplete guidance can lead to certain actions on the part of the student and can foster their curiosity, thus providing them with learning opportunities. For example, a response like "24 x 10 = ?" does not provide guidance, yet it is actionable. This further demonstrates the need to treat the dimensions as independent. In terms of the other qualities of the {\tt Expert} responses, they do not normally reveal the answer and tend to include scaffolding; however, there are a small number of instances where they failed to identify the mistake or its location. Overall, we conclude that human responses from {\tt Expert} are significantly better than {\tt Novice}.

\paragraph{\textit{Tutor tone} and \textit{Human-likeness}:} Our findings on the \textit{Tutor Tone} align with those of \citet{wang2024bridging} -- in task-oriented conversations, AI tutors tend to be more \textit{Neutral} than \textit{Encouraging}. When we combine these two labels into "Non-offensive", the DAMR score reaches 100\% as we observe no offensive responses from any LLMs or humans. We observe high scores for most of the LLMs on \textit{human-likeness}, which demonstrates their capability to generate human-like output with minimal or no grammatical and fluency mistakes, showing the timely nature of our study, which focuses more on in-depth semantic and pedagogical aspects of tutor responses rather than only on superficial attributes like grammaticality and fluency. 

\paragraph{Tutor response quality on \texttt{Bridge} vs. \texttt{MathDial}:} As discussed in Section \ref{sec:benchprep}, the conversational contexts in the {\tt Bridge} dataset are typically very short (see Table \ref{tab:datastats}) and the dialogues are grounded in elementary math operations, so most models are able to identify the mistakes and their locations. However, they struggle to provide appropriate guidance without revealing the answer because the mistakes are generally related to quite basic operations like addition or multiplication, often in a one-step type of mathematical problems. Still, models like \texttt{GPT-4} and \texttt{Llama-3.1-405B} are able to offer some reasonable guidance. In contrast, for \texttt{MathDial}, the contexts are longer, the mistakes are grounded in reasoning, and the responses are more structured. Yet, many LLMs do not meet the expectations for each dimension of the taxonomy, as discussed earlier. DAMR scores for {\tt Bridge} and {\tt Mathdial} are shown in Appendix Table \ref{tab:bridge_results} and \ref{tab:mathdial_results}, respectively. Combining both types of data in {\tt MRBench} makes it both challenging and comprehensive.

\paragraph{Overall performance:} 
In summary, all LLMs and even human tutors lack some pedagogical abilities required for effective tutoring. While \texttt{Llama-3.1-405B} is the most effective, followed by {\tt Mistral} and other state-of-the-art models, {\tt GPT-4} reveals the answer too quickly. {\tt Gemini} is less coherent and accurate, and {\tt Sonnet} focuses on human-likeness and encouraging tone but is less effective in other dimensions. {\tt Phi3} is the worst-performing model according to our analysis, as it fails to understand the context, while \texttt{Llama-3.1-8B}, despite being smaller, performs reasonably well. Human responses are also not perfect -- {\tt Novice} responses are ambiguous and short, whereas {\tt Expert} responses are more focused on actionability and less on other dimensions. Overall, the proposed taxonomy precisely categorizes performance across 8 dimensions, reflecting the current state-of-the-art in AI tutors. Our study demonstrates that there is a considerable room for improvement in the pedagogical abilities of AI tutors.

\paragraph{Reliability of LLM-based Evaluation:}

We also performed annotations using {\tt Prometheus2} and {\tt Llama-3.1-8B} as \textit{critic} LLMs. The correlation scores with human annotations are presented in Appendix Tables \ref{tab:corr_results1} and \ref{tab:corr_results2}, respectively. Across both LLMs, it can be observed that most of the correlation scores are negative (except for the human-likeness dimension), indicating that the annotations from the LLMs are unreliable for the challenging pedagogical dimensions. There may be several reasons for this: (i) {\tt Prometheus2} is not trained on our taxonomy dimensions, except for the general human-likeness dimension, where the model shows slightly better correlations with positive scores. However, the score for human-likeness remains low and requires gold-standard responses, which are not unique and were unavailable in our case. (ii) We believe both LLMs have a limited understanding of rich pedagogical concepts, as they were not specifically trained on pedagogically rich datasets.

At the same time, we acknowledge that the experiments presented in this work are preliminary and have several limitations, including reliance on a specific prompt (see Figure \ref{fig:evalprompt}) and the use of only two LLMs. Therefore, it is possible that better results may be achieved via extensive prompt engineering and experimentation with other LLMs as critics. We leave such experiments to future work.

 \vspace{-0.1cm}
\section{Conclusion}
\label{sec:con}
 \vspace{-0.2cm}

This paper presents the first effort to unify AI tutor evaluation for the student mistake remediation task in the mathematics domain. Specifically, we propose an evaluation taxonomy with eight pedagogical dimensions based on the key learning sciences principles. We also release the {\tt MRBench} benchmark with seven state-of-the-art LLM-as-tutors responses, along with gold human annotations. We discuss the limitations of each LLM by pinpointing the lack of specific pedagogical abilities demonstrated in their responses based on human evaluation. We also assess the feasibility of LLMs as evaluators in this context by correlating their judgements with human annotations, indicating that they are often unreliable. The study demonstrates that the current state-of-the-art LLMs are not yet sufficiently good as AI tutors and there is a huge scope for improvement, also identifying the most relevant directions for such improvement. We hope that the resources released with this study will streamline the evaluation process and help track the progress in the development of AI tutors. Furthermore, our study opens possibilities for creation and annotation of datasets that can be used for RLHF and fine-tuning, helping future AI tutors align with human and pedagogical values. 

\section*{Limitations}
\label{sec:limit}

We believe that this study provides a useful starting point for  streamlining the evaluation of AI tutors. However, we acknowledge that there are certain limitations of this work, and addressing these limitations in the future is an important task.

\paragraph{Establishing the relationships between evaluation dimensions} This study evaluates tutor response quality across the proposed eight dimensions independently. However, in practice, these dimensions may be inherently interrelated and may influence one another. A comprehensive investigation of these interdependencies can facilitate more effective modeling and ranking of tutor responses according to their quality at the dialogue level. The annotations provided in {\tt MRBench} serve as a foundational resource for future research in this direction. 

\paragraph{Extensions beyond the task of student mistake remediation and to subjects other than mathematics} 
The proposed taxonomy primarily focuses on the task of the student mistake remediation in the domain of mathematics. We acknowledge that the proposed taxonomy will need to be verified on and likely adapted if applied to other tasks such as concept learning, and to subjects other than mathematics. However, we believe that the proposed taxonomy, grounded in the learning sciences principles, will provide useful guidelines for future research.

\paragraph{Taking the students' perspective into account} The current taxonomy and annotation scheme focus on the appropriateness of the tutor responses. However, one of the limitations is that it does not consider the tutoring dialogues' impact on the overall student learning. Specifically, the annotation pertains to the individual tutor turns within educational dialogues, which restricts our understanding of broader implications on student learning processes and learning gains, typically observed after a conversation concludes. We believe that the atomic tutor response evaluation at the utterance level, as presented in this study, should in the future be scaled up to the conversation level to better assess AI tutors' pedagogical abilities. 

\paragraph{Evaluation with other LLMs as critics}
In this study, we limit the LLM-based evaluation to two LLMs as critics, using the evaluation prompt presented in Figure \ref{fig:evalprompt}. The results obtained with these LLMs are not encouraging, as detailed in Section \ref{sec:result&discc}. Future research should explore state-of-the-art and more powerful LLMs as critics and experiment with diverse prompt templates. At the same time, we believe that this preliminary study provides a basis and a benchmark for further investigation.

\section*{Ethics Statement}
\label{sec:ethics}

Although we do not foresee any ethical risks, we acknowledge that this work relies on the outputs from LLMs, and there are certain risks associated with such outputs in general since these models may generate responses that, although plausible, can be factually incorrect, nonsensical, or even offensive. Of particular importance for the educational domain is the fact that hallucinations can misguide students and propagate biases. Nevertheless, we strongly believe that this study will help shed light on the current capabilities of LLMs in the context of educational dialogues, and the insights gained from this study may help mitigate issues related to the use of LLMs in the educational domain in the future.

\section*{Acknowledgments}
\label{sec:ack}
This research is partially supported by Google through the Google Academic Research Award (GARA) 2024. We are grateful for their support. We also extend our gratitude to the campus supercomputing center at MBZUAI.

\bibliography{custom}
\appendix

\section{Evaluation Taxonomy, Annotation Labels, and Desiderata}
The definitions, associated labels, and the desired labels for each dimension of the proposed taxonomy are provided in Table \ref{tab:det_def}.

\paragraph{Completeness of the evaluation taxonomy:} Through an iterative analysis of the taxonomy, we identify eight dimensions that comprehensively assess tutor response quality in the context of mistake remediation. However, other educational settings, particularly those involving tutorial dialogues beyond mistake remediation, may require modifications, as discussed in the limitations section. To establish a robust framework, we initially considered additional dimensions such as grammaticality and empathy, among others. However, our\textit{ validation pilot study} (see Section \ref{sec:valideevaltax}) confirmed that the selected eight dimensions are both necessary and sufficient for evaluating tutor response quality in dialogues aimed at mistake remediation.

\begin{table*}[!htb]
    \centering
    \resizebox{\textwidth}{!}{
    \begin{tabular}{|l|l|l|c|}
         \hline
        \textbf{Dimension} &  \textbf{Definition} & \textbf{Labels} & \textbf{Desiderata} \\ \hline
        Mistake identification & Has the tutor identified/recognized a mistake in a student’s response? & \makecell[l]{(1) Yes \\ (2) To some extent \\ (3) No} & Yes \\ \hline
        Mistake location & Does the tutor’s response accurately point to a genuine mistake and its location? & \makecell[l]{(1) Yes \\ (2) To some extent \\ (3) No} & Yes \\ \hline
        Revealing of the answer & Does the tutor reveal the final answer (whether correct or not)? & \makecell[l]{(1) Yes (and the revealed answer is correct) \\ (2) Yes (but the revealed answer is incorrect) \\ (3) No} & No\\ \hline
        *Providing guidance & \makecell[l]{Does the tutor offer correct and relevant guidance, such as an explanation, \\ elaboration, hint, examples, and so on?} & \makecell[l]{(1) Yes \\ (2) To some extent \\ (3) No} & Yes \\ \hline
        Actionability & Is it clear from the tutor’s feedback what the student should do next? & \makecell[l]{(1) Yes \\ (2) To some extent \\ (3) No} & Yes \\ \hline
        Coherence & Is the tutor’s response logically consistent with the student’s previous responses? & \makecell[l]{(1) Yes \\ (2) To some extent \\ (3) No} & Yes \\ \hline
        Tutor tone & Is the tutor’s response encouraging, neutral, or offensive? & \makecell[l]{(1) Encouraging \\ (2) Neutral \\ (3) Offensive} & Encouraging \\ \hline
        Human-likeness & Does the tutor’s response sound natural rather than robotic or artificial? & \makecell[l]{(1) Yes \\ (2) To some extent \\ (3) No} & Yes \\ \hline
    \end{tabular}}
    \caption{\small An overview of the proposed evaluation taxonomy, including associated annotation labels and desired expected labels. *For the guidance dimension, we  provide further details for the labels: `Yes' indicates that guidance is correct and relevant to the mistake; `To some extent' indicates that guidance is provided but is either partially/fully incorrect or incomplete; and `No' indicates that no guidance has been provided.}
    \label{tab:det_def}
\end{table*}

\begin{table*}[!t]
    \centering
    \resizebox{\textwidth}{!}{ 
    \begin{tabular}{@{}lcccccccc@{}}
        \hline \hline
        \textbf{Tutor}      & \textbf{Mistake Identification} & \textbf{Mistake Location} & \textbf{Revealing of the Answer} & \textbf{Providing Guidance} & \textbf{Actionability} & \textbf{Coherence}  & \textbf{Tutor Tone} & \textbf{Human-likeness} \\ \hline \hline
        *{\tt Novice}       & -0.37 & 0.09 & -0.56 & -0.72 & 0.15 & -0.15 & -0.71 & 0.18 \\ 
        {\tt Expert}       & -0.01 & -0.25 & -0.13 & -0.19 & -0.08 & -0.11 & -0.40 & 0.01 \\\hline
        {\tt Phi3}         & -0.67 & -0.58 & -0.51 & -0.51 & -0.46 & -0.33 & -0.62 & 0.03 \\
        {\tt Llama-3.1-8B}    & -0.12 & -0.37 & -0.17 & 0.04 & -0.07 & -0.16 & -0.29 & 0.11 \\\hline
        {\tt Gemini}       & 0.02 & 0.09 & -0.06 & -0.16 & -0.12 & -0.07 & -0.24 & 0.07 \\
        {\tt Sonnet}       & -0.11 & -0.12 & -0.21 & -0.11 & -0.22 & -0.08 & -0.2 & 0.07 \\
        {\tt Mistral}      & -0.06 & -0.11 & -0.10 & -0.23 & -0.15 & -0.20 & -0.19 & 0.06 \\
        \hline
        {\tt GPT-4}        & -0.07 & 0.01 & -0.20 & -0.21 & 0.02 & -0.02 & -0.11 & 0.08 \\
        {\tt Llama-3.1-405B}  & -0.03 & -0.08 & -0.05 & -0.05 & 0.00 & 0.06 & -0.13 & 0.11 \\ \hline

        \hline
    \end{tabular}}
    \caption{Annotation correlation (AC) scores between human annotations and judgments from \texttt{Prometheus2} as LLM critic across different tutors and evaluation dimensions on {\tt MRBench}. The correlation scores are calculated using \textit{Pearson’s correlation} \cite{sedgwick2012pearson}. *Only 60 dialogues were considered for {\tt Novice}, whereas all 192 dialogues were considered for {\tt Expert} and other tutors.}
    \label{tab:corr_results1}
\end{table*}

\begin{table*}[!t]
    \centering
    \resizebox{\textwidth}{!}{ 
    \begin{tabular}{@{}lcccccccc@{}}
        \hline \hline
        \textbf{Tutor}      & \textbf{Mistake Identification} & \textbf{Mistake Location} & \textbf{Revealing of the Answer} & \textbf{Providing Guidance} & \textbf{Actionability} & \textbf{Coherence}  & \textbf{Tutor Tone} & \textbf{Human-likeness} \\ \hline \hline
         *{\tt Novice}       & -0.42 & 0.06 & -0.71 & -0.80 & 0.17 & -0.17 & -0.77 & 0.14 \\ 
        {\tt Expert}       & -0.03 & -0.29 & -0.17 & -0.23 & -0.10 & -0.16 & -0.49 & -0.01 \\\hline
        {\tt Phi3}         & -0.71 & -0.67 & -0.77 & -0.73 & -0.61 & -0.41 & -0.62 & 0.04 \\
        {\tt Llama-3.1-8B}    & -0.08 & -0.46 & -0.17 & 0.09 & -0.09 & -0.23 & -0.38 & 0.09 \\\hline
        {\tt Gemini}       & 0.06 & 0.12 & -0.11 & -0.27 & -0.22 & -0.09 & -0.34 & 0.09 \\
        {\tt Sonnet}       & -0.07 & -0.17 & -0.26 & -0.21 & -0.29 & -0.08 & -0.32 & 0.07 \\
        {\tt Mistral}      & -0.16 & -0.16 & -0.16 & -0.34 & -0.27 & -0.28 & -0.17 & 0.07 \\
        \hline
        {\tt GPT-4}        & -0.03 & 0.01 & -0.13 & -0.23 & 0.01 & -0.05 & -0.06 & 0.10 \\
        {\tt Llama-3.1-405B}  & -0.01 & -0.02 & -0.01 & -0.07 & 0.00 & 0.02 & -0.06 & 0.09 \\ \hline
        \hline
    \end{tabular}}
    \caption{Annotation correlation (AC) scores between human annotations and judgments from \texttt{Llama-3.1-8B} as LLM critic across different tutors and evaluation dimensions on {\tt MRBench}. The correlation scores are calculated using \textit{Pearson’s correlation} \cite{sedgwick2012pearson}. *Only 60 dialogues were considered for {\tt Novice}, whereas all 192 dialogues were considered for {\tt Expert} and other tutors.}
    \label{tab:corr_results2}
\end{table*}

\section{Prompt Template for Generating LLM Responses}
\label{appsec:prompttempgenresp}

\begin{figure*}[!htb]
    \centering
    \begin{subfigure}[b]{0.68\linewidth}
        \centering
        \includegraphics[width=\linewidth]{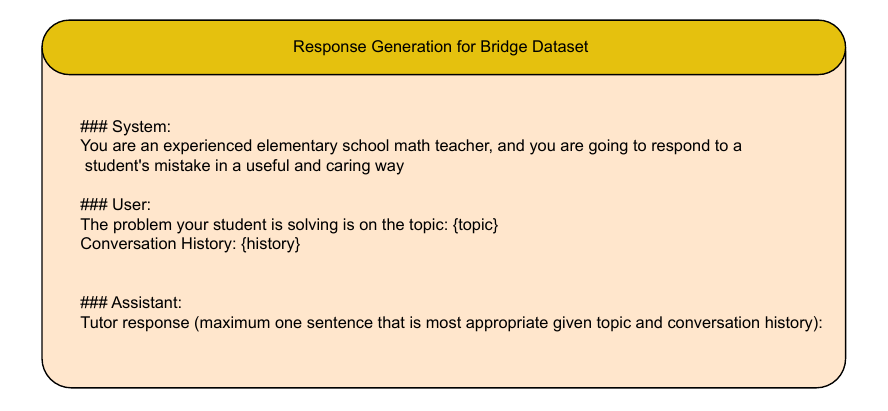}
    \end{subfigure}
    \hfill
    \begin{subfigure}[b]{0.68\linewidth}
        \centering
        \includegraphics[width=\linewidth]{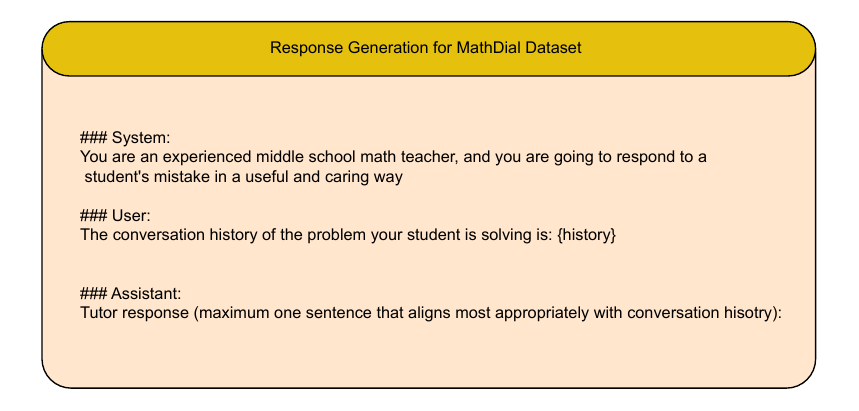}
    \end{subfigure}
    \caption{The prompt template used to generate responses from the seven considered LLMs for both {\tt Bridge} and {\tt MathDial} datasets. The template is adapted from \citet{wang2024bridging}. The notable differences are: (1) The problems covered in the {\tt Bridge} dataset are at the elementary school level, whereas those in {\tt MathDial} are at the middle school level; and (2) The conversation topic is not provided in {\tt MathDial}.}
    \label{fig:promptresponse}
\end{figure*}

The prompt template used to generate responses from the seven considered LLMs for both the {\tt Bridge} and {\tt MathDial} datasets is shown in Figure \ref{fig:promptresponse}. The template is adapted from \citet{wang2024bridging}.

\newpage
\section{Human Annotators Training}
\label{appsec:humann}
\begin{figure*}[!htb]
    \centering
    \includegraphics[width=0.7\linewidth]{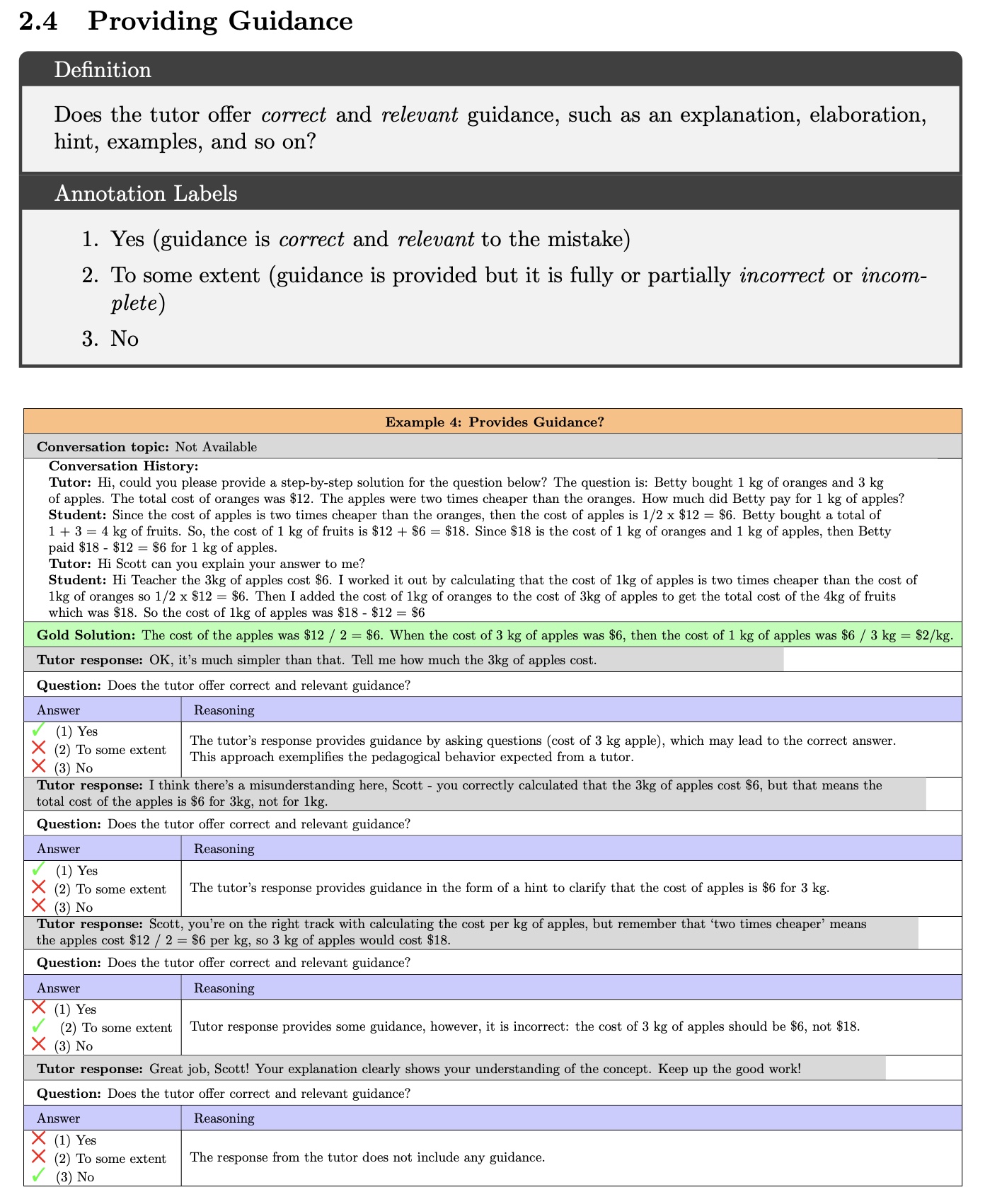}
    \caption{An example from the guidelines document provided to annotators, showing a page that details definitions, annotation labels, and associated examples for the \textit{Providing Guidance} dimension of the taxonomy.}
    \label{fig:guiddoc}
\end{figure*}

\begin{figure}[!htb]
    \centering
    \begin{subfigure}[b]{0.6\linewidth}
        \centering
        \includegraphics[width=\linewidth]{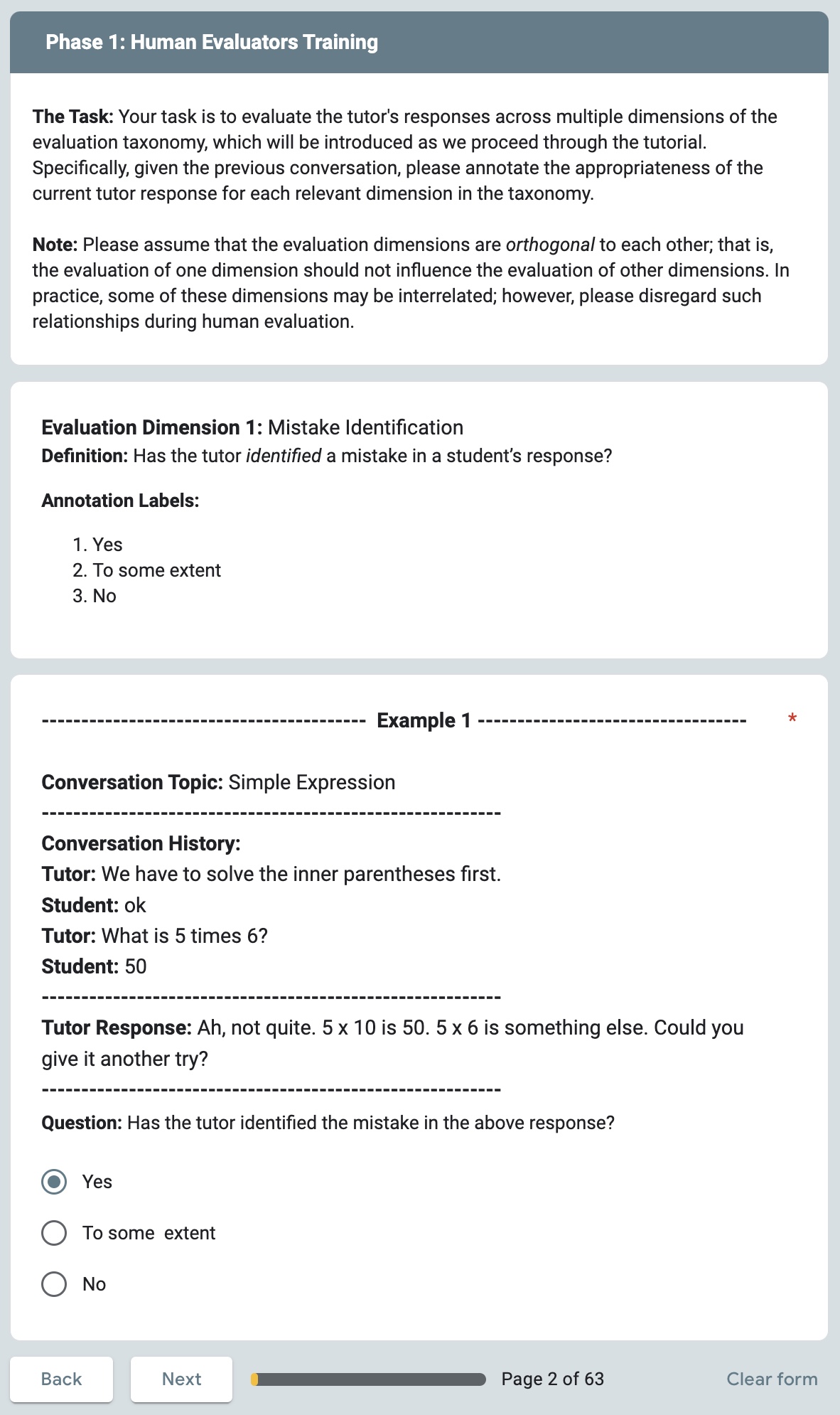}
        \caption{Training Question}
    \end{subfigure}
    \hfill
    \begin{subfigure}[b]{0.6\linewidth}
        \centering
        \includegraphics[width=\linewidth]{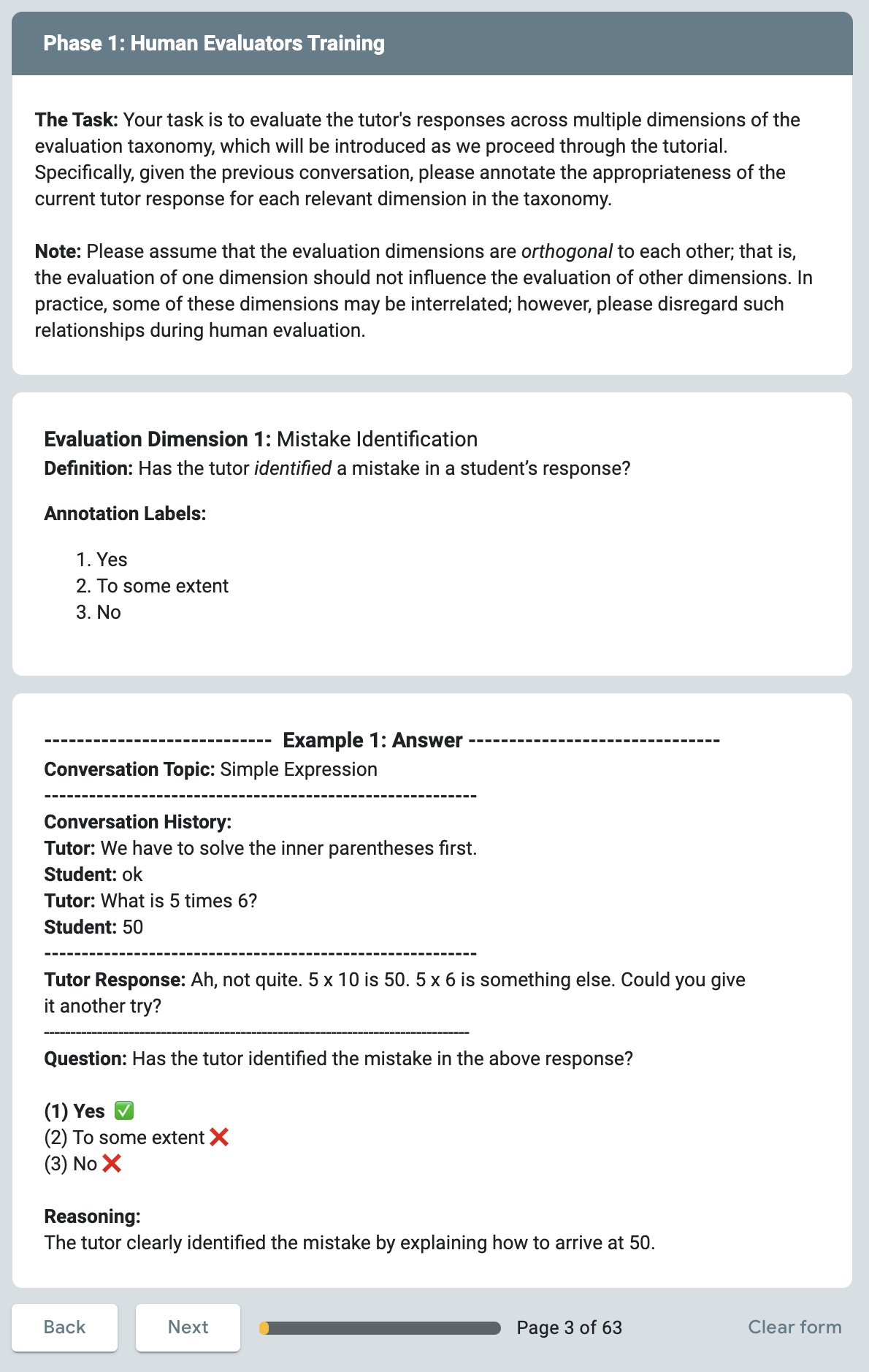}
        \caption{Feedback}
    \end{subfigure}
     \caption{An example from the annotator training phase for the \textit{Mistake Identification} dimension.}
    \label{fig:sctrain}
\end{figure}

\begin{figure}[!htb]
    \centering
    \begin{subfigure}[b]{0.6\linewidth}
        \centering
        \includegraphics[width=\linewidth]{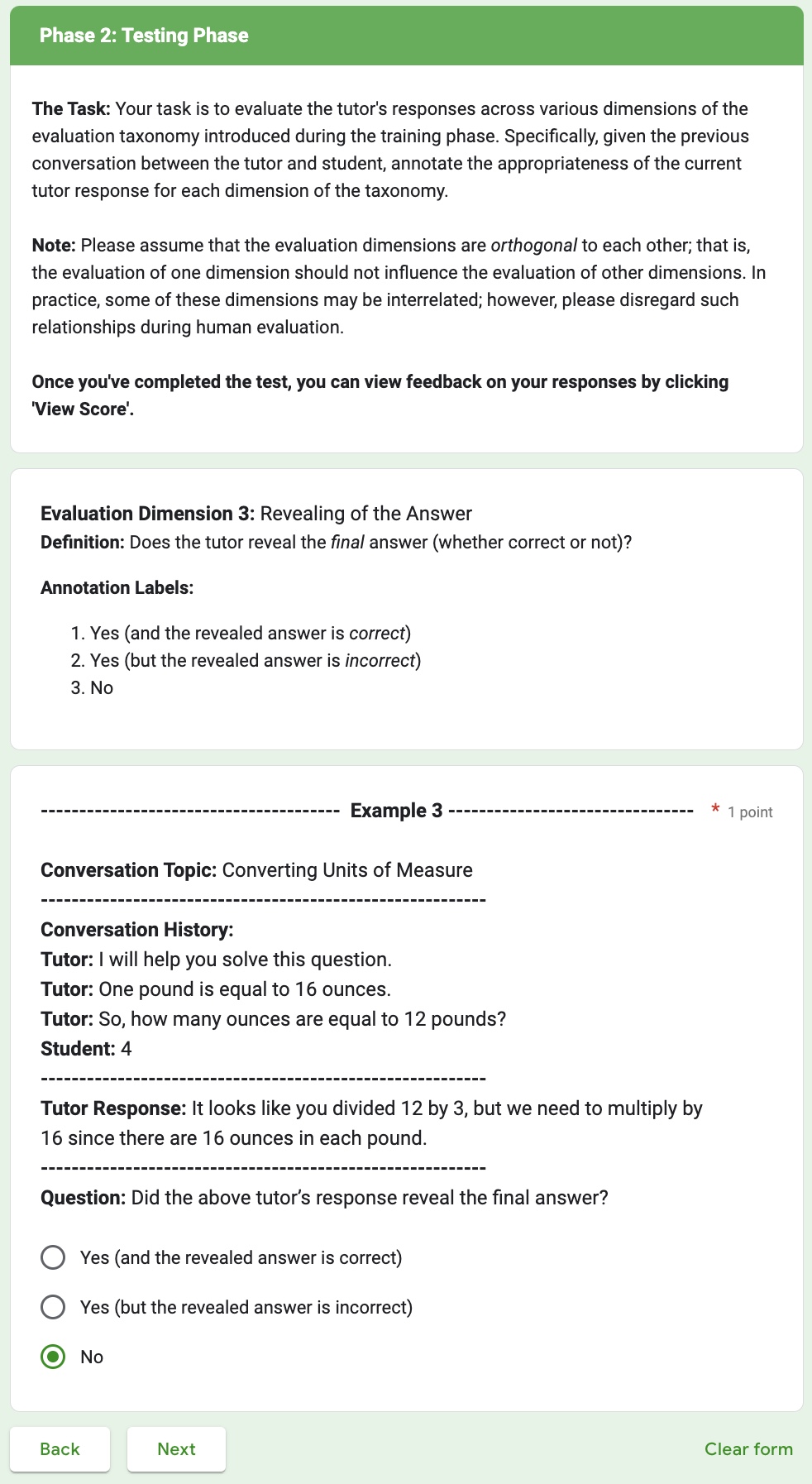}
        \caption{Testing Question}
    \end{subfigure}
    \hfill
    \begin{subfigure}[b]{0.6\linewidth}
        \centering
        \includegraphics[width=\linewidth]{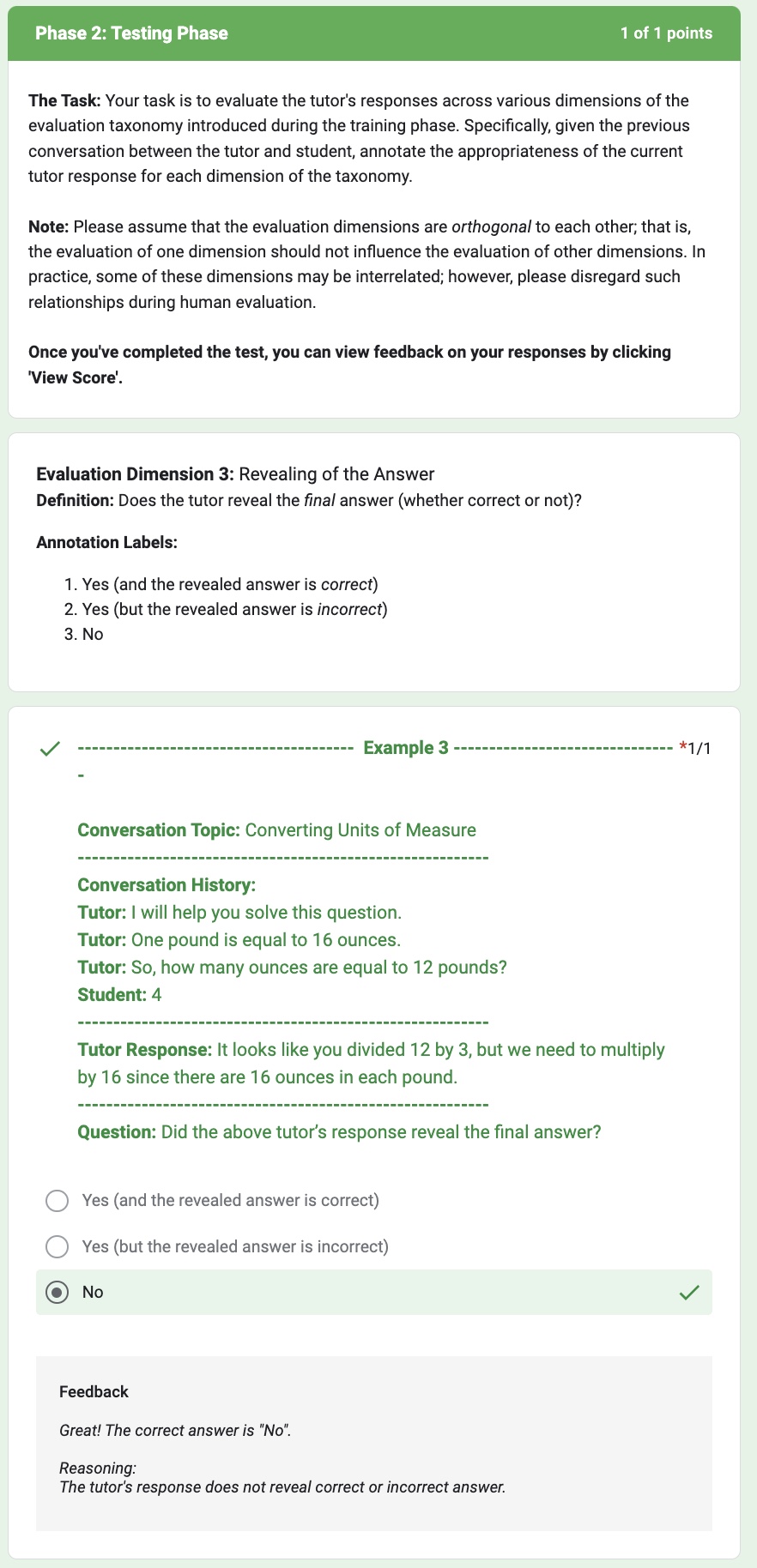}
        \caption{Feedback}
    \end{subfigure}
     \caption{An example from the annotator testing phase for the \textit{Revealing of the Answer} dimension.}
    \label{fig:sctest}
\end{figure}

As discussed in Section \ref{sec:humananno}, prior to commencing large-scale human annotation, we implemented a two-phase interactive training and evaluation protocol and asked each annotator to undertake training. A representative screenshot from the interactive training phase is provided in Figure \ref{fig:sctrain}. Subsequently, we assessed annotators' understanding through a structured quiz, as is shown in a screenshot presented in Figure \ref{fig:sctest}. Additionally, we developed a comprehensive set of annotation guidelines, serving as a reference for annotators during the large-scale annotation process. An example from the guidelines document is shown in Figure \ref{fig:guiddoc}.

\section{Benchmark Statistics}
\label{appsec:benchstats}

Table \ref{tab:datastats} shows the statistics for the {\tt Bridge}, {\tt MathDial}, and {\tt MRBench} datasets. It can be observed that the conversation history and response lengths from different LLMs and humans are generally shorter in the {\tt Bridge} dataset compared to the {\tt MathDial} dataset. Additionally, the number of turns differs between them. These aspects highlight that including both datasets in {\tt MRBench} ensures diversity and provides for a good mix of easy and difficult mathematical problems, making the benchmark both comprehensive and challenging.

\begin{table}[!htb]
    \centering
    \resizebox{0.48\textwidth}{!}{
    \begin{tabular}{l|c|c|c}
        \hline \hline 
        \textbf{Parameters} & {\tt \textbf{Bridge}} & {\tt \textbf{MathDial}} & {\tt \textbf{MRBench}} \\ \hline \hline 
        % Add your data rows below
        \#Dialogues & 60 & 132 & 192 \\ 
         Avg. turns & 4.00 & 5.51 & 5.04 \\ 
         Avg. dialogue length & 140.59 & 906.20 & 1247.25 \\ \hline
        \#LLM responses &  420 & 924 &  1344 \\
        \#Human responses &  120 & 132 &  152\\
        \#Total responses &  540 & 1056 &  1596 \\
        \#Total annotations* &  540$\times$8 & 1056$\times$8 &  1596$\times$8 \\
        \hline 
        Avg. {\tt Novice} response length & 
        45.31 & - &  45.31 \\ 
        Avg. {\tt Expert} response length &
        75.38 &  89.13 &  85.01 \\
        Avg. {\tt Phi3} response length & 
        128.85 & 273.96 & 231.30 \\
        Avg. {\tt Llama-3.1-8B} response length &
        157.68 & 223.88 & 204.97 \\
        Avg. {\tt Gemini} response length & 
        106.57 & 144.87 & 139.08 \\
        Avg. {\tt Sonnet} response length &
        111.22 & 160.69 &  146.63\\
        Avg. {\tt Mistral} response length &
        93.01 & 148.98 & 133.07 \\
        Avg. {\tt GPT-4} response length &
        118.59 & 229.87 & 198.24\\
        Avg. {\tt Llama-3.1-405B} response length &
        163.81 & 225.13 & 229.04 \\ \hline
        \#Humans as tutors & 2 & 1 & 2 \\
        \#LLMs  as tutors & 7 & 7 & 7 \\
        \hline
         \hline 
    \end{tabular}}
    \caption{Dataset statistics for {\tt Bridge}, {\tt MathDial}, and {\tt MRBench}. * indicates that the annotations are considered for 8 evaluation dimensions of the taxonomy. In all cases, length is estimated using the \textit{number of characters}.}
    \label{tab:datastats}
\end{table}

\newpage
\section{Prompt Template for LLM-based  Evaluation}
\label{appsec:prompttempllmannt}

\begin{figure*}[!htb]
    \centering
    \includegraphics[width=0.7\linewidth]{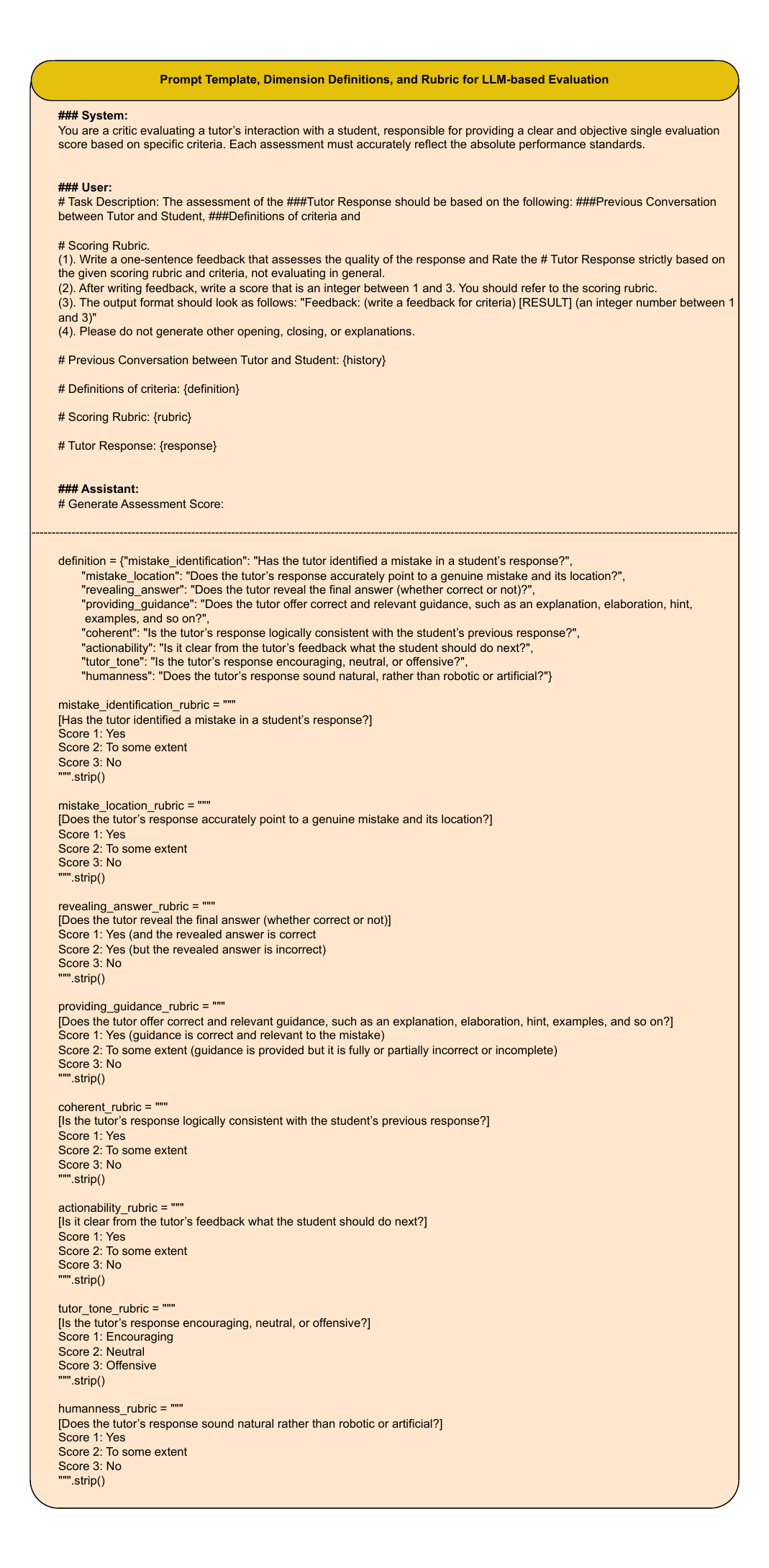}
    \caption{Prompt template for evaluation with LLMs as critics \cite{kim2024prometheus}.}
    \label{fig:evalprompt}
\end{figure*}

Figure \ref{fig:evalprompt} illustrates the prompt template we have adapted for the evaluation with LLMs as critics \cite{kim2024prometheus}. The template is based on the insights drawn from the {\tt Prometheus2} model's official guidelines.\footnote{\url{https://github.com/prometheus-eval/prometheus-eval}}

\begin{table*}[h!]
\centering
\resizebox{\textwidth}{!}{
\begin{tabular}{lcccccccc}
\hline \hline
\textbf{Tutor}         & \textbf{Mistake Identification} & \textbf{Mistake Location} & \textbf{Revealing of the Answer} & \textbf{Providing Guidance} & \textbf{Coherence} & \textbf{Actionability} & \textbf{Tutor Tone} & \textbf{Human-likeness} \\ \hline \hline
\texttt{Novice} & 43.33 & 16.67 & 80.00 & 11.67 & 1.67 & 50.00 & 90.00 & 35.00 \\
\texttt{Expert} & 82.44 & 76.20 & 89.52 & 68.36 & \textbf{71.88} & 88.06 & 14.43 & 86.77 \\ \hline
\texttt{Llama-3.1-8B} & 85.23 & 73.57 & 43.49 & 52.22 & 51.66 & 87.13 & 28.44 & 93.27 \\
\texttt{Phi3} & 78.02 & 74.22 & 43.49 & 55.18 & 29.29 & 79.62 & 39.59 & 82.79 \\ \hline
\texttt{Gemini} & 67.40 & 56.63 & 38.19 & 54.81 & 48.39 & 67.79 & 31.01 & 11.29 \\
\texttt{Sonnet} & 89.82 & 85.32 & \textbf{91.40} & 61.38 & 65.56 & 96.06 & \textbf{51.10} & \textbf{95.07} \\
\texttt{Mistral} & 93.52 & 83.27 & 78.08 & 64.42 & 59.49 & 90.50 & 14.35 & 94.03 \\ \hline
\texttt{GPT-4} & \textbf{98.74} & \textbf{93.91} & 25.66 & 74.08 & 63.53 & 96.84 & 46.65 & 83.68 \\
\texttt{Llama-3.1-405B} & \textbf{98.74} & \textbf{93.91} & 62.47 & \textbf{79.28} & 62.07 & \textbf{97.13} & 18.31 & 93.02 \\ \hline \hline

\end{tabular}}
\caption{Pedagogical ability assessment of different LLMs using the DAMR scores (in \%) across eight evaluation dimensions with \textit{human evaluation} on the {\tt Bridge} data. The best DAMR scores for each dimension are \textbf{bolded.}}
\label{tab:bridge_results}
\end{table*}

\begin{table*}[h!]
\centering
\resizebox{\textwidth}{!}{
\begin{tabular}{lcccccccc}
\hline \hline
\textbf{}         & \textbf{Mistake Identification} & \textbf{Mistake Location} & \textbf{Revealing of the Answer} & \textbf{Providing Guidance} & \textbf{Coherence} & \textbf{Actionability} & \textbf{Tutor Tone} & \textbf{Human-likeness} \\ \hline \hline
\texttt{Novice}   & -                               & -                         & -                                & -                          & -                 & -                      & -                   & -                  \\ 
\texttt{Expert} & 73.13 & 57.03 & 91.12 & 66.66 & \textbf{77.93} & 75.13 & 11.17 & 87.83 \\ \hline
\texttt{Llama-3.1-8B} & 77.93 & 46.11 & 87.81 & 42.17 & 38.64 & 77.82 & 15.86 & 93.97 \\
\texttt{Phi3} & 6.21 & 4.14 & 87.81 & 0.68 & 4.11 & 21.38 & 47.91 & 38.12 \\ \hline
\texttt{Gemini} & 61.03 & 31.83 & 81.13 & 29.63 & 40.13 & 51.76 & 17.73 & 94.11 \\
\texttt{Sonnet} & 83.42 & 62.73 &\textbf{96.33} & 58.47 & 58.84 & 85.12 & \textbf{56.32} & \textbf{96.93} \\
\texttt{Mistral} & 93.10 & 68.97 & 90.27 & 63.14 & 75.23 & 85.38 & 15.44 & 95.89 \\ \hline
\texttt{GPT-4} & \textbf{92.24} & \textbf{80.05} & 65.60 & \textbf{76.93} & 38.54 & 87.14 & 33.34 & 92.32 \\
\texttt{Llama-3.1-405B} & \textbf{92.24} & \textbf{80.05} & 89.03 & 76.08 & 80.12 & 89.19 & 15.17 & 89.53 \\ \hline \hline
\end{tabular}}
\caption{Pedagogical ability assessment of different LLMs using the DAMR scores (in \%) across eight evaluation dimensions with \textit{human evaluation} on the {\tt MathDial} data. `-' indicates that DAMR scores for {\tt Novice} are not available for {\tt MathDial} data. The best DAMR scores for each dimension are \textbf{bolded.}}
\label{tab:mathdial_results}
\end{table*}

\end{document}